# Uncertainty-Aware Calibration of a Hot-Wire Anemometer With Gaussian Process Regression

Rubén A. García-Ruiz, José L. Blanco-Claraco  , Javier López-Martínez,
and Ángel J. Callejón-Ferre

*Abstract*—**Expensive ultrasonic anemometers are usually required to measure wind speed accurately. The aim of this work is to overcome the loss of accuracy of a low cost hot-wire anemometer caused by the changes of air temperature, by means of a probabilistic calibration using Gaussian Process Regression. Gaussian Process Regression is a non-parametric, Bayesian, and supervised learning method designed to make predictions of an unknown target variable as a function of one or more known input variables. Our approach is validated against real datasets, obtaining a good performance in inferring the actual wind speed values. By performing, before its real use in the field, a calibration of the hot-wire anemometer taking into account air temperature, permits that the wind speed can be estimated for the typical range of ambient temperatures, including a grounded uncertainty estimation for each speed measure.**

## I. INTRODUCTION

**H**OT-WIRE sensors are low-cost devices usually employed to measure wind speed, and sometimes the speed of other fluids. They comprise a thin metallic wire with a typical diameter in the range 0.5-5 $\mu$m, and a length of 1 mm. They are usually made of platinum, tungsten, or platinum-iridium.

Their operating principle consists in heating the wire with an electric current (Joule effect) up to some temperature above the ambient and then exposing it to the incident fluid flow such that it is cooled by, mainly, convective heat transfer. The fluid velocity can then be inferred as a function of the heat transfer from the heated wire and the fluid. Hot-wire anemometers can be classified, depending on their control architecture, into: constant-temperature anemometer (CTA), constant-current anemometer (CCA), and constant-voltage anemometer (CVA). The difference between them depends on the variable whose set-point is the input of the control circuitry, namely, resistance temperature, electric current, or applied voltage, respectively.

Hot-wire anemometers have been used for decades in a wide range of applications that require measuring the speed of a fluid [1]–[4]. In particular, they are well suited for low-flow rate measurements, and manufacturers often recommend its use for low to medium wind speeds. As will be seen in section III.C, a good performance has been observed for speeds up to 20 m/s, and we would not recommend using this kind of sensors for higher speeds. The reason is twofold: (a) the error and the uncertainty of the prediction would increase, and (b) due to the mechanical fragility of the sensor. Although a minimum detectable velocity is not provided by the manufacturer of the sensor at test in this work, this research found that small changes in the range 0.1-0.2 m/sare resolvable.

Hot-wire anemometers are nowadays widely-used for their high measuring bandwidth, which allows detecting fast velocity fluctuations. Their small size and low weight also make them suitable for applications with limited space. They are easy to handle, low cost and additionally, they require very little power to operate, enabling their use in battery-powered devices [5].

Calibration of hot-wire anemometers is typically carried out for some predefined *constant temperature*. This becomes one of the main disadvantages of this type of sensors [6]: if they operate inside a fluid flow at a different temperature than the one used during the calibration, measurements will not be accurate. Some authors have developed different methods for correcting wind speed measurements in hot-wire anemometers [7]–[10]. These methods typically require other application-specific parameters such that the kinematic viscosity and thermal conductivity of the fluid. However, air temperature has a significant influence in wind speed corrections [11], [12].

Most practical applications of wind speed sensing imply operating at temperatures that vary through the day and the different seasons in the year. Even if the sensor works isolated or covered, the temperature might still present significant variations. It is well-known that readings from hot-wire anemometers depend on both, the ambient and the wire temperature [7]–[10]. For this reason, we need to consider ambient temperature as an extra variable to obtain wind speed from their non-linear relationship (generally, wire temperature is always considered).

The present work proposes using machine learning (ML) techniques to approach this calibration problem. In particular, we show how *Gaussian Process Regression* (GPR) [13] has the best performance from all the methods included in our comparison. A *Gaussian Process* (GP) is a distribution over functions, and GPR is a non-parametric, Bayesian, supervised learning method, with wide applications in the industry and academic research [14]–[16]. In a nutshell, GPR takes a set of samples and builds a model from them by estimating the posterior joint probability of the GP, hence building a model able to make predictions about values not observed in the samples. A key characteristic of GP is its capability of providing a *measure of uncertainty* for each prediction. Also, a GPR can express any prior knowledge, e.g. from a human expert, by means of *a priori* probability density functions. It has a good adaptability in dealing with complex non-linear problems with small samples.

In comparison to other non-linear, widely-used machine learning methods such as Support Vector Machines (SVM) [17]–[20] or Artificial Neural Networks (ANN) [21]–[23], GPR has the advantages of being easier to implement, self-adaptive to enable superior parameter estimation, flexible enough to make non-parametric inferences [24], and providing a grounded estimation for the output uncertainty. We claim that the latter is of paramount importance for any engineering process, since any physical measurement, direct or indirect, should be accompanied by its expected accuracy.

The main contribution of the present work is two-fold: (i) we discuss and justify what metrics should be observed to decide among different competing regression techniques in order to select the one with the best predictive performance, and (ii) we apply such methodology to the study of how a low-cost hot-wire anemometer can be calibrated by means of machine learning techniques to overcome its most important drawback, namely its loss of accuracy when air temperature changes. The result is the identification of a method that enables the use of low-cost anemometers with reasonable accuracy within a typical range of ambient temperatures, therefore enabling its use in a wide range of applications where the low cost of these devices might be a significant advantage, for example, in large networks of sensors.

This paper is organized as follows. First, the theoretical bases of GPR are introduced in Section II. The experimental setup and the metrics used for the evaluation of the model are detailed in Section III, together with the experimental results and its discussion. Finally, some conclusions are drawn in Section IV.

## II. BACKGROUND

This section first provides a brief summary of common regression techniques, including Gaussian Process Regression (GPR), and then introduces the basis of GPR for the particular setup employed in this work with a greater detail, given the importance of this method in subsequent experimental results.

### A. Regression Methods

Regression is the problem of finding a suitable model to predict the values of one or more dependent variables (outputs) given the known values of the independent variables (inputs). Each one of the existing regression models typically has a small number of *parameters* which must be *learned* or *fitted* from training data: pairings of input and output variables.

Next, we enumerate the different regression methods included in our comparison (refer to Table II), as named in their reference implementation from MATLAB's *Statistics and Machine Learning Toolbox* (SMLT). In-depth reviews on each technique can be found elsewhere in the vast related bibliography [25]–[29].

*Linear regression models* [27] are easy to interpret and fast to evaluate, but often lack a precise predictive accuracy.

*Regression trees* are non-parametric models which naturally define subgroups, scale well with the complexity of the data, and are not limited by the number of predictor variables [30]. They are easy to interpret, fast for fitting and predicting, and have a reduced memory cost.

*Support Vector Machine* (SVM) regression is a nonparametric technique, relying on kernel functions, where data are mapped into a high dimensional feature space via nonlinear mapping, after which a linear regression is performed in this feature space [25]. Linear SVMs are easy to interpret, but may have low predictive accuracy, while nonlinear SVMs are more difficult to interpret, but can be more accurate. The SVM regressions compared in this work are:

- Linear: the kernel function is linear. The model flexibility is low.
- Quadratic: the kernel function is quadratic. The model flexibility is medium.
- Cubic: the kernel function is cubic. The flexibility of the model is medium.
- Fine, medium, and coarse: These models are the same except for different Kernel scale values of $\sqrt{P}/4$, $\sqrt{P}$, and $4\sqrt{P}$, respectively, with $P$ the number of predictors. The response function of "fine" is well-suited for rapid variations, while "coarse" better fits very slowly-varying signals.

*Ensembles of Trees* combine several regression trees to achieve better predictive performance than the corresponding single regression trees [31]. The following versions are compared:

- Boosted Trees: it consists in least-squares boosting with regression tree learners. The model flexibility is medium-high.
- Bagged Trees: it consists in bootstrap aggregating or bagging, with regression tree learners. The flexibility of this model is high.

*Gaussian Process Regression* provides a probabilistic model on the space of functions, as discussed in Section II-B. MATLAB's toolbox implementation automatically fits the method flexibility to offer a small error while simultaneously protecting against overfitting. Kernel functions that are frequently used in the literature are: Rational Quadratic, Squared Exponential, Matern 5/2, and Exponential.

### B. Background on Gaussian Process Regression

Consider a training data set $\mathbf{D}$ of $n$ observations, $\mathbf{D} = \{(\mathbf{x}_i, y_i) | i = 1, \ldots, n\}$, where $\mathbf{x}$ is an input vector of dimension $N$, and $y$ is a scalar output or target. Given a new input $\mathbf{x}^*$ (test input), the goal of the regression is to obtain the predictive distributions that have not been seen in the training set. On the basis of training data, the aim is to obtain a function that makes predictions for all possible input values. To carry out this, assumptions about the characteristics of the underlying function must be made, as otherwise any function which is consistent with the training data would be equally valid.

GPs can be seen as a generalization of the Gaussian probability distribution to a distribution over *functions*. A GP performs inference directly in the space of functions, giving a prior probability to each possible function (where higher probabilities are given to functions that are considered to be more likely) and learning the target function from the training data.

The specification of the prior is important, because it fixes the properties of the functions considered for inference. These properties are entirely dictated by the covariance function, which is symmetric and positive semi-definite for any input point $\mathbf{x}$. The covariance function specifies the covariance between two or more random variables and, typically, the covariance functions have a number of free parameters called hyperparameters. Finding suitable hyperparameters for the covariance function is the biggest problem of learning in GP. The hyperparameters give us a model of the data and characteristics (such as smoothness, length-scale and stationary) which we can interpret. Thus, the covariance function is the most important factor in order to control the properties of a GP, thus it must be carefully selected [11], [32]. In our study, we will use different covariance functions and we will compare them in order to choose the one achieving the best predictive performance for our training data set.

A GP over a function (to be estimated) $f : \mathbb{R}^N \to \mathbb{R}$ is entirely specified by its mean function, $m(\mathbf{x})$, and a covariance function, $\mathbf{k}(\mathbf{x}, \mathbf{x}')$, for any two points of the state space $\mathbf{x}, \mathbf{x}' \in \mathbb{R}^N$, such that:

$$m(\mathbf{x}) = \mathbb{E}[f(\mathbf{x})] \tag{1}$$

$$\begin{aligned} \mathbf{k}(\mathbf{x}, \mathbf{x}') &= cov(f(\mathbf{x}), f(\mathbf{x}')) \\ &= \mathbb{E}[(f(\mathbf{x}) - m(\mathbf{x}))(f(\mathbf{x}') - m(x'))] \end{aligned} \tag{2}$$

and the GP itself is denoted as:

$$f(\mathbf{x}) \sim GP(m(\mathbf{x}), \mathbf{k}(\mathbf{x}, \mathbf{x}')) \tag{3}$$

where we used $a \sim b$ to denote "$a$ follows the probability distribution $b$". In practice, we should also take into account the *noise*, which is customarily assumed to be an additive, independent identically distributed (i.i.d.) Gaussian noise $\varepsilon$ with zero mean and variance $\sigma_n^2$, that is:

$$y = f(\mathbf{x}) + \varepsilon, \quad \varepsilon \sim \mathcal{N}(0, \sigma_n^2) \tag{4}$$

and where $\mathcal{N}(\cdot, \cdot)$ denotes the multivariate Gaussian or normal distribution with the given mean and covariance matrix. For the sake of simplicity in notation, the mean function is usually taken to be zero and we will consider it in this way (note that the mean of the posterior process is not confined to be zero). Then, the prior distribution of the observation target $y$ is:

$$y \sim \mathcal{N}(0, \mathbf{K}(\mathbf{X}, \mathbf{X}) + \sigma_n^2 \mathbf{I}_n), \quad \text{with } \mathbf{K}(\mathbf{X}, \mathbf{X}) = (K_{ij})_{n \times n} \tag{5}$$

where $\mathbf{X}$ denotes the $n \times N$ matrix of the $n$ training samples of dimensionality $N$, $K(\cdot, \cdot)$ refers to the matrix with the entries given by the covariance function $k(\cdot, \cdot)$ being the matrix elements $K_{ij} = k(x_i, x_j)$ and $\mathbf{I}_n$ is the $n$-dimensional identity matrix.

The joint probability distribution of the training and test sets according to the definition of GP follow a Gaussian distribution. Then, the joint distribution of the observed target values ($\mathbf{y}$) and the test function at new inputs values ($f_*$) is:

$$\begin{bmatrix} \mathbf{y} \\ f_* \end{bmatrix} \sim \mathcal{N} \left( \mathbf{0}_{n \times 1}, \begin{bmatrix} \mathbf{K}(\mathbf{X}, \mathbf{X}) + \sigma_n^2 \mathbf{I}_n & \mathbf{K}(\mathbf{X}, \mathbf{X}_*) \\ \mathbf{K}(\mathbf{X}_*, \mathbf{X}) & \mathbf{K}(\mathbf{X}_*, \mathbf{X}_*) \end{bmatrix} \right) \tag{6}$$

where, in a similar way to $X$, $X_*$ is defined as the $n \times N$ matrix of the $n$ testing $N$-length input vectors.

Conditioning the prior to the observed training outputs and taking into account that the posterior distribution over functions is also a Gaussian, we obtain the key predictive equations for GPR:

$$\begin{aligned} \overline{f_*} &\triangleq \mathbb{E}[f_* | \mathbf{X}, \mathbf{y}, \mathbf{X}_*] \\ &= \mathbf{K}(\mathbf{X}_*, \mathbf{X})[\mathbf{K}(\mathbf{X}, \mathbf{X}) + \sigma_n^2 \mathbf{I}_n]^{-1} \mathbf{y} \end{aligned} \tag{7}$$

$$\begin{aligned} \mathbf{cov}(f_*) &= \mathbf{K}(\mathbf{X}_*, \mathbf{X}_*) - \mathbf{K} \\ & \cdot (\mathbf{X}_*, \mathbf{X})[\mathbf{K}(\mathbf{X}, \mathbf{X}) + \sigma_n^2 \mathbf{I}_n]^{-1} \mathbf{K}(\mathbf{X}, \mathbf{X}_*) \end{aligned} \tag{8}$$

where $\overline{f_*}$ and $\mathbf{cov}(f_*)$ are the estimated mean of the predictive distribution and its covariance matrix, respectively. Therefore, given a test sample, and on the basis of the training set and covariance function, a GPR model can predict a mean value $\overline{f_*}$ (our best estimation for $\mathbf{y}_*$) and a variance which represents the uncertainty of the estimated output.

### C. Hyperparameters Selection

The selection of hyperparameters, together with the choice for the covariance function, are the key design factors of GPR. Selection of optimal hyperparameters, stacked in the vector $\boldsymbol{\theta}$, is done by maximizing the marginal likelihood function $p(\mathbf{y}|\mathbf{X}, \boldsymbol{\theta})$. However, following the common practice, due to its superior numerical stability and efficiency, the corresponding negative log likelihood is minimized instead:

$$\log p(\mathbf{y}|\mathbf{X}, \boldsymbol{\theta}) = -\frac{1}{2}\mathbf{y}^T \mathbf{K}_y^{-1} \mathbf{y} - \frac{1}{2}\log |\mathbf{K}_y| - \frac{n}{2}\log 2\pi \tag{9}$$

where $\mathbf{K}_y = \mathbf{K}(\mathbf{X}, \mathbf{X}) + \sigma_n^2 \mathbf{I}_n$ is the covariance matrix for the output vector $\mathbf{y}$. The negative log marginal likelihood is a statistical technique used for estimating the optimal parameters of a model. The first term $(-\frac{1}{2}\mathbf{y}^T \mathbf{K}_y^{-1} \mathbf{y})$ measures how well the model fits the data, the second term $(-\frac{1}{2}\log |\mathbf{K}_y|)$ is a complexity penalization term and the third term $(-\frac{n}{2}\log 2\pi)$ is a normalization constant.

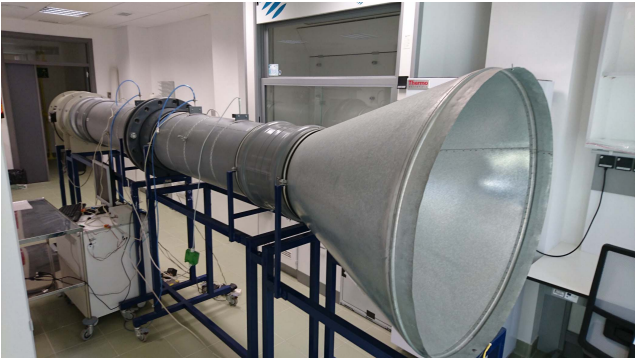

(a)

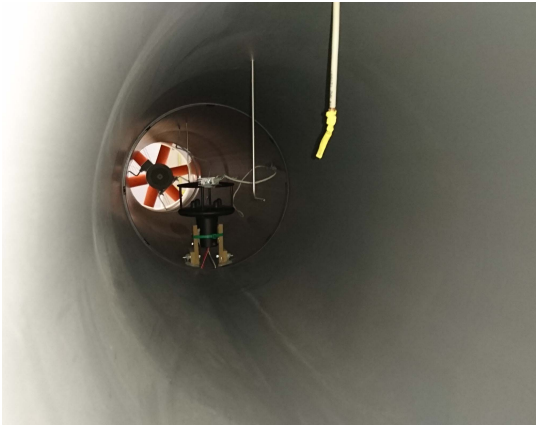

(b)

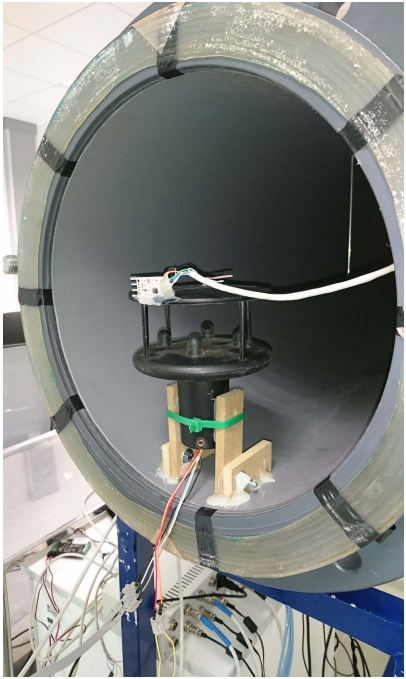

(c)

Fig. 1. Wind tunnel used in this work. (a) Overview of the entire system. (b) View of inside of the tunnel, where the temperature and wind sensors are installed. (c) Detail view of the placing of sensors, with the tunnel split in two during the experiment preparation.

To find out the minimum of Eq. (9) and, consequently to find the optimal hyperparameters, the conjugate gradient method is usually used. The conjugate gradient is an iterative method of optimizing functions based on gradient-ascent. The calculation of the partial derivatives of the log marginal likelihood with respect to the hyperparameters is required [13]:

$$\frac{\partial}{\partial \boldsymbol{\theta}_i} \log p(\mathbf{y}|\mathbf{X}, \boldsymbol{\theta}) = \frac{1}{2} \mathbf{y}^T \mathbf{K}_y^{-1} \frac{\partial \mathbf{K}_y}{\partial \boldsymbol{\theta}_i} \mathbf{K}_y^{-1} \mathbf{y} \quad (10)$$

$$- \frac{1}{2} tr(\mathbf{K}_y^{-1} \frac{\partial \mathbf{K}_y}{\partial \boldsymbol{\theta}_i})$$

$$= \frac{1}{2} tr((\boldsymbol{\alpha}\boldsymbol{\alpha}^T - \mathbf{K}_y^{-1}) \frac{\partial \mathbf{K}_y}{\partial \boldsymbol{\theta}_i}) \quad (11)$$

where $tr(\cdot)$ is the trace of its matrix argument, and $\boldsymbol{\alpha} = \mathbf{K}_y^{-1}\mathbf{y}$.

One of the main drawbacks of GPR is the complexity of computing the marginal likelihood because of the matrix inverse operation. In theory, it has a complexity of $\mathcal{O}(n^3)$ with $n$ the dimension of training inputs. Although efficient factorizations can be applied (e.g. Cholesky) instead of naive matrix inversion, this operation still remains as a bottleneck of the training procedure.

## III. Case Study

### A. Experimental Setup

Experiments were carried out inside a wind tunnel at the University of Almería (Spain). The tunnel has a length of 4.74 m, a circular cross-section of 38.8 cm diameter, contraction ratio of 1:5:32 and the coefficient between the entrance diameter and the length of the contraction section is 0.92 [33], [34]. An axial fan (Model HCT-45-2T-3/AL, Sodeca S.A., Sant Quirze of Besora, Spain) induces the air flow in the wind tunnel, and a Micromaster 420 Inverter (Siemens Energy & Automation Inc., Alpharetta, GA, USA) is used to control the fan speed, by modulating the current frequency between 0 and 50 Hz.

The anemometer under calibration is a hot-wire anemometer of the popular model "revision C" by "Modern Device". To perform the calibration we rely on a more reliable anemometer, model "Windsonic" by Gill Instruments Ltd, as ground truth. The latter is an ultrasonic anemometer, with a measurement range of 0 to 60 m/s and a precision of ± 2%. In addition, air temperature is measured by means of a PT100 in order to improve the accuracy of the hot-wire anemometer own temperature measurements.

In order to achieve the calibration of the wind sensor, it is necessary to find the relationship (if it exists) between the inputs and outputs, such that the calibrated model predicts outputs as close as possible to the real values. The inputs of our system are considered to be the raw voltage readings from the hot-wire anemometer and the air temperature from the PT100 sensor, while the output of the system is the wind speed measured from the ultrasonic anemometer. Data of these three sensors (voltage of hot-wire anemometer in volts, air temperature in Celsius degrees and wind of ultrasonic anemometer in meters per second) were measured every 2 seconds while wind speed was varied between 0 and 21 m/s. The controller allows changing the speed continuously (e.g. a velocity ramp) but speed was increased step by step instead, in order to allow the flow inside the tunnel to stabilize. We waited 20 seconds after each speed change to ensure that both the sensor and the flow were stable before picking a measure for the dataset. Several campaigns of measurements were performed at different temperatures from 19 to 30 °C.

The measurement cycle was similar for all temperatures at test: starting at 0 m/s, the wind speed was increased in small steps, while attempting to provide a good sampling of low to medium wind speeds, where hot-wire anemometers are more reliable and find their most common working conditions. Therefore, our experiments mainly focus on wind speed values up to 10-15 m/s, approximately. On the other hand, wind speed was increased modulating the current frequency manually and waiting for a determined time to stabilization and then for another time period to allow enough data records to be grabbed. These periods were measured manually, hence the existence of more data in some measure cycles. Overall, more than 4000 input-output data points were obtained.

### B. Evaluation of the Model

To assess the performance of the model, different metrics have been used:

- *Mean Absolute Error (MAE):* it measures the average of all absolute errors between predictions and ground truth values. It reveals how similar the predicted values are to the ground truth values. $MAE = \frac{1}{n}\sum_{i=1}^{n} |y_i - \hat{y}_i|$, where $y_i$ is the ground truth value of the i-th sample, and $\hat{y}_i$ is the corresponding predicted value.

- *Root Mean Square Error (RMSE):* it measures the square root of the average of all squared absolute errors between predictions and ground truth values. It reveals the overall deviation of both values. $RMSE = \sqrt{\frac{1}{n}\sum_{i=1}^{n} |y_i - \hat{y}_i|^2}$, where $y_i$ is the ground truth value of the i-th sample, and $\hat{y}_i$ is the corresponding predicted value.

Models having low MAE and RMSE are preferred. Both metrics evaluate the model prediction error and are indifferent to the sign of error. The main differences between them is that RMSE gives a relatively high weight to large errors, since the errors are squared before they are averaged. For that reason, RMSE is preferred when large errors are particularly undesirable.

- *Coefficient of determination or $R^2$:* it provides a measure of how well future samples are likely to be predicted by the model. The value of $R^2$ always lies between $-1$ and $+1$. Values close to zero represent no association between the variables, whereas values close to $-1$ or $1$ indicate strong relationship between predictions and ground truth values.

$$R^2 = 1 - \frac{\sum_{i=1}^{n} (y_i - \hat{y}_i)^2}{\sum_{i=1}^{n} (y_i - \overline{y})^2}, \text{ where: } \overline{y} = \frac{1}{n}\sum_{i=1}^{n} y_i \quad (12)$$

where $\overline{y}$ is the average of the ground truth values, $y_i$ is the ground truth value of the i-th sample and $\hat{y}_i$ is the corresponding predicted value.

### C. Experimental Results and Discussion

Next, we expose the experimental validation of the proposed GPR-based wind speed estimator. The flow chart of the calibration process can be seen in figure 2. As it can be observed in figure 2, the entire training set has been used to train the GPR model: both the matrix $X$, corresponding to hot-wire anemometer voltage and air temperature, and the vector $Y$ corresponding to the real wind speed measured with the ultrasonic anemometer. Once the model has been trained, the matrix $X_*$ of the test set (hot-wire anemometer voltage and air temperature) is used to predict wind speed and obtain the corresponding confidence interval. Finally, the vector $Y_*$ (real wind speed) of the test set is used to analyze the error committed and evaluate the GPR model.

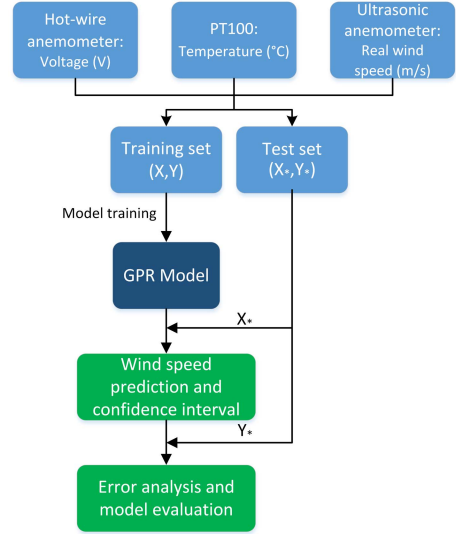

Fig. 2. Flow chart describing the proposed calibration and validation process.

|  | **Traning set** | **Test set** |
|---|---|---|
| Section III.C.1 | Cross Validation with 6 folds (4112 points) | |
| Section III.C.2 | 4112 points | |
| Section III.C.3 | 70% (2878 points) | 30% (1234 points) |
| Section III.C.4a | 3583 points | 529 points |
| Section III.C.4b | 2691 points | 1421 points |
| Section III.C.4c | 3319 points | 793 points |
| Section III.C.4d | 3554 points | 558 points |
| Section III.C.5 | 70% (2878 points) | 30% (1234 points) |

Different sizes of data are included in the training and test sets. In Table I are summarized the number of points selected for the training and test sets in the GPR model for each subsection of Section III.C.

To ensure a correct estimation of credible intervals,[1] the GPR model needs to account for the additive Gaussian noise employed in the model [16]. MATLAB's toolbox for Gaussian process models optimizes the standard deviation of that noise, denoted as "Sigma", while training from a given input data set. Finally, when the GPR model makes a prediction, it also generates a prediction interval by considering the uncertainty of both, the additive noise (the "Sigma" value),

---

[1] Although in most Engineering literature the term used is "confidence interval", according to [35], [36] in the Bayesian Statistics "credible interval" is a more accurate term.

TABLE II

Evaluation of Different Regression Models for Our Wind Speed Dataset. (LR: Linear Regression; RT: Regression Tree; SVM: Support Vector Machine; ET: Ensembles of Trees; GPR: Gaussian Process Regression)

| Model | MAE (m/s) | RMSE (m/s) | $R^2$ |
|---|---|---|---|
| LR: Linear | 1.4243 | 1.8351 | 0.82 |
| LR: Interactions Linear | 1.4238 | 1.8354 | 0.82 |
| LR: Robust Linear | 1.3403 | 1.9453 | 0.80 |
| LR: Stepwise Linear | 1.4243 | 1.8351 | 0.82 |
| RT: Fine Tree | 0.1843 | 0.3304 | 0.99 |
| RT: Medium Tree | 0.2384 | 0.4086 | 0.99 |
| RT: Coarse Tree | 0.4091 | 0.6579 | 0.98 |
| SVM: Linear | 1.3240 | 1.9231 | 0.80 |
| SVM: Quadratic | 0.6526 | 0.8154 | 0.96 |
| SVM: Cubic | 3.7074 | 5.4092 | -0.55 |
| SVM: Fine Gaussian | 0.3327 | 0.4224 | 0.99 |
| SVM: Medium Gaussian | 0.3899 | 0.5156 | 0.99 |
| SVM: Coarse Gaussian | 0.5761 | 0.7449 | 0.97 |
| ET: Boosted Trees | 0.4181 | 0.5974 | 0.98 |
| ET: Bagged Trees | 0.2349 | 0.3682 | 0.99 |
| GPR: Rational Quadratic | 0.1699 | 0.2820 | **1.00** |
| GPR: Squared Exponential | 0.1964 | 0.3122 | 0.99 |
| GPR: Matern 5/2 | 0.1808 | 0.2918 | **1.00** |
| GPR: Exponential | **0.1582** | **0.2776** | **1.00** |

TABLE III

Comparison Between the Different GPR Models at Test, Using an Information-Based Criterion (BIC)

| Kernel function | Log marginal likelihood (L) | Kernel parameters (m) | Number of data points (n) | BIC |
|---|---|---|---|---|
| Rational Quadratic | -868,5 | 3 | 4112 | 1748.9 |
| Squared Exponential | -1115,1 | 2 | 4112 | 2237.5 |
| Matern 5/2 | -964,7 | 2 | 4112 | 1936.7 |
| Exponential | **-755,4** | 2 | 4112 | **1518.1** |

TABLE IV

Results of MAE, RMSE and $R^2$ for Training and Test Points, of the GPR Model With Exponential as Covariance Function, Using Randomly-Selected 70% of All Data Points for Training and the Other 30% for Testing

| | MAE (m/s) | RMSE (m/s) | $R^2$ |
|---|---|---|---|
| Training | 0.0958 | 0.1718 | 0.99846 |
| Test | 0.1620 | 0.2833 | 0.99563 |

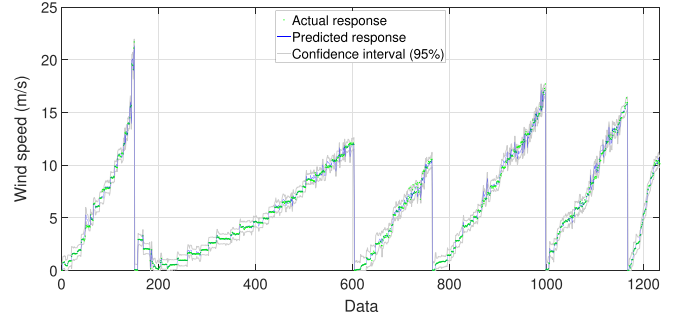

Fig. 3. Predictions, 95% credible interval of the GPR model and real wind values for the test points. Each individual ramp represents data from a run at a different ambient temperature.

and the uncertainty value of the parameters learned from the data.

*1) Comparison of Regression Methods:* First, it is convenient to assess whether GPR is the best technique for the data under study, in terms of being able to make accurate predictions. To verify this, we trained multiple regression models and evaluated their "validation" errors. These process was carried out with MATLAB's regression Learner App, included in the Statistics and Machine Learning Toolbox. The entire data set was used to training the models, and Cross Validation was used with 6 folds. Folds can be understood as subsets of data. Cross Validation partitions the data in folds, trains the model using the out-of-fold observations, assesses the model performance using in-fold data and finally calculates the average test error over all folds. This method makes an efficient use of all the data and permits to obtain a good estimation of the predictive accuracy of the final model. The resulting errors of the regression models are shown in Table II.

The results confirm that Gaussian Process Regression is the regression model that best fits the data. In particular, Gaussian Process Regression with exponential as covariance function produce the lowest MAE and RMSE, and a good value of $R^2$.

*2) Information-Based Comparison of GPR Models:* In order to compare the GPR models, the Bayesian Information Criterion (BIC) [37], [38] has been used. BIC is a metric based on the on the highest posterior probability to finding the best model for make predictions. BIC model is defined as:

$$BIC = -2\ L + m \log n \qquad (13)$$

where $L$ denotes the log marginal likelihood, $p$ the number of Kernel parameters and $n$ the number of data points employed in the model. The model fits better the data when lower value of BIC is obtained. The likelihood takes into account both, how close the predicted values are to ground-truth, and how large is the predicted uncertainty.

Table III shows the BIC obtained for each GPR model, where it is clear that GPR with exponential as covariance function is the best one, in accordance with the metrics MAE, RMSE, and $R^2$ discussed above.

*3) Cross Validation of GPR: Method 1:* Gaussian Process Regression with exponential covariance function is applied to the data. Training points are selected randomly and constitute the 70% of the full data, whereas test points are the remaining 30%. The training and test sets are normalized in [0,1] and the optimal hyperparameters are obtained through the conjugate gradient method. The predictions and the variance are calculated with the training and test sets using Matlab2017b. The results of the GPR model are summarized in Table IV, where it can be seen the MAE, RMSE and $R^2$ values for both training and test points. MAE and RMSE for test points are 0.1620 m/s and 0.2833 m/s, respectively, while $R^2$ for these test points is also high (0.99563); ultimately, errors in the prediction are therefore small.

In Figure 3 are presented the predictions of the GPR model for the test points, along with the estimation 95% credible interval and the ground truth values corresponding to the ultrasonic anemometer. As it can be observed, the predicted values are close to the real ones and the real values in almost

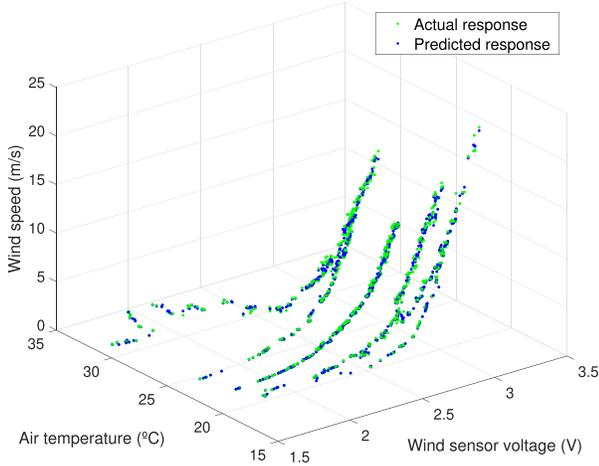

Fig. 4. Predictions and real wind values for the test points, plotted as raw voltage output from the hot-wire anemometer vs. air temperature. Each filament-like cluster of data points represents a run at a different room temperature.

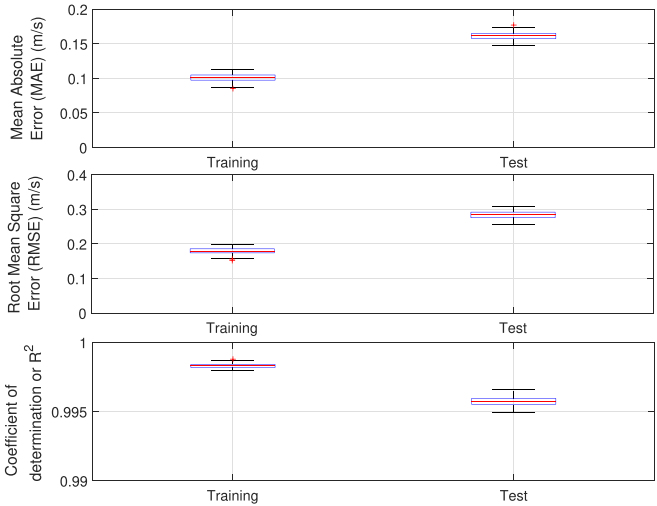

Fig. 5. Boxplot of MAE, RMSE, and $R^2$ of the GPR model for 100 iterations with different training (70% of all data set points) and test sets (the remaining 30%). As expected, in any cross-validation test the performance obtained for the training set is better than that for the test subset of the data.

all the cases fall within the credible interval, which indicates a high accuracy in the GPR model.

In Figure 4, the same predictions and ground truth values are shown in 3D along with the voltage of the hot-wire anemometer and the air temperature. It can be observed the strong relationship between both variables and also the accuracy of the predictions.

The results are obviously influenced by the training data, which are selected randomly. It could be thought that with other training data, worse results would be obtained. To evaluate it, 100 iterations have been done selecting randomly different training sets and consequently test sets, always complying that 70% of the data is used for training and the remaining 30% for test. The results of the evaluation of the model are shown in a boxplot in Figure 5. On each box, the red line is the median, the box edges are the 25th and 75th percentiles, the whiskers include until the most extreme data points not considering outliers, which are represented individually as red crosses. Similar MAE, RMSE and $R^2$ are obtained, which indicates that the model gives a good approximation of the real wind as function of the hot-wire anemometer voltage and air temperature.

*4) Cross Validation of GPR: Method 2:* As an alternative cross validation of GPR to predict wind speed, we now propose to select training and testing sets, not as a given fraction of the overall data set, as done in the previous section, but selecting entire dataset runs for some given temperature values. In this case the GPR is trained *without* any single observation of the sensor response for some specific temperature, and we will evaluate its performance in *inferring* ("interpolating") its behavior from the response at other temperatures.

Part of the results are shown in Figure 6. In this case, the average RMSE of all cases is 0.024 m/s for the training datasets and 1.734 m/s for the testing datasets. The average MAE is 0.012 m/s and 1.373 m/s for the training and testing datasets, respectively. As expected, these values are similar to the results in Table IV for the training parts, but much higher for the test datasets. This could be explained by the lack of information the GPR has to make predictions about the sensor behavior in conditions it has not been able to learn from. However, it is remarkable that the probabilistic nature of GPR allows to have a predicted uncertainty for each prediction, and in most cases where the error is large, uncertainty is high as well –refer to Figure 6. In particular, notice how disallowing the GPR to learn the sensor behavior for one of the extreme temperatures included in our study (the dataset for 30°C), leads to the largest errors, since the estimator in this case is extrapolating, not interpolating, the sensor behavior for those conditions. To quantify and demonstrate this fact, we evaluated the average RMSE (1.51 m/s) and MAE (1.13 m/s) when predictions are "interpolated". On the other hand, the average RMSE and MAE values of the two datasets in which predictions are "extrapolated" are 2.305 m/s and 1.968 m/s, respectively, validating the insight that predictions are less accurate when they need to be extrapolated.

*5) Sensibility to Ground-Truth Errors:* Since our model proposes using the *actual* air temperature as an input to the wind speed sensor model, it is in order wondering how much does the air temperature measurement of the sensor affects the results. Our experimental set up employs a PT100 for air temperature measurement, with an accuracy of $\pm\,0.06\,°C$ at $0\,°C$. To assess its influence in the performance of the GPR model, different errors in the ambient temperature measurement have been introduced. First, GPR model have been trained with the real values of air temperature measured with the PT100. Next, we have introduced two different errors in the test datasets and have evaluated the predictions:

- *Random error:* four levels of random noise have been evaluated in order to simulate different degrees of sensor accuracy. We considered the scenarios of air temperature accuracies of: $\pm\,0.1\,°C$, $\pm\,0.2\,°C$, $\pm\,0.5\,°C$, and $\pm1\,°C$, respectively.
- *Systematic error:* an example of a typical systematic error might be not protecting the temperature sensor from direct solar radiation, which strongly affects its measurements. Four levels of systematic error have been evaluated: $+ 0.25 \ °C$, $+ 0.5 \ °C$, $+ 1 \ °C$, and $+ 1.5 \ °C$, with respect to the real value.

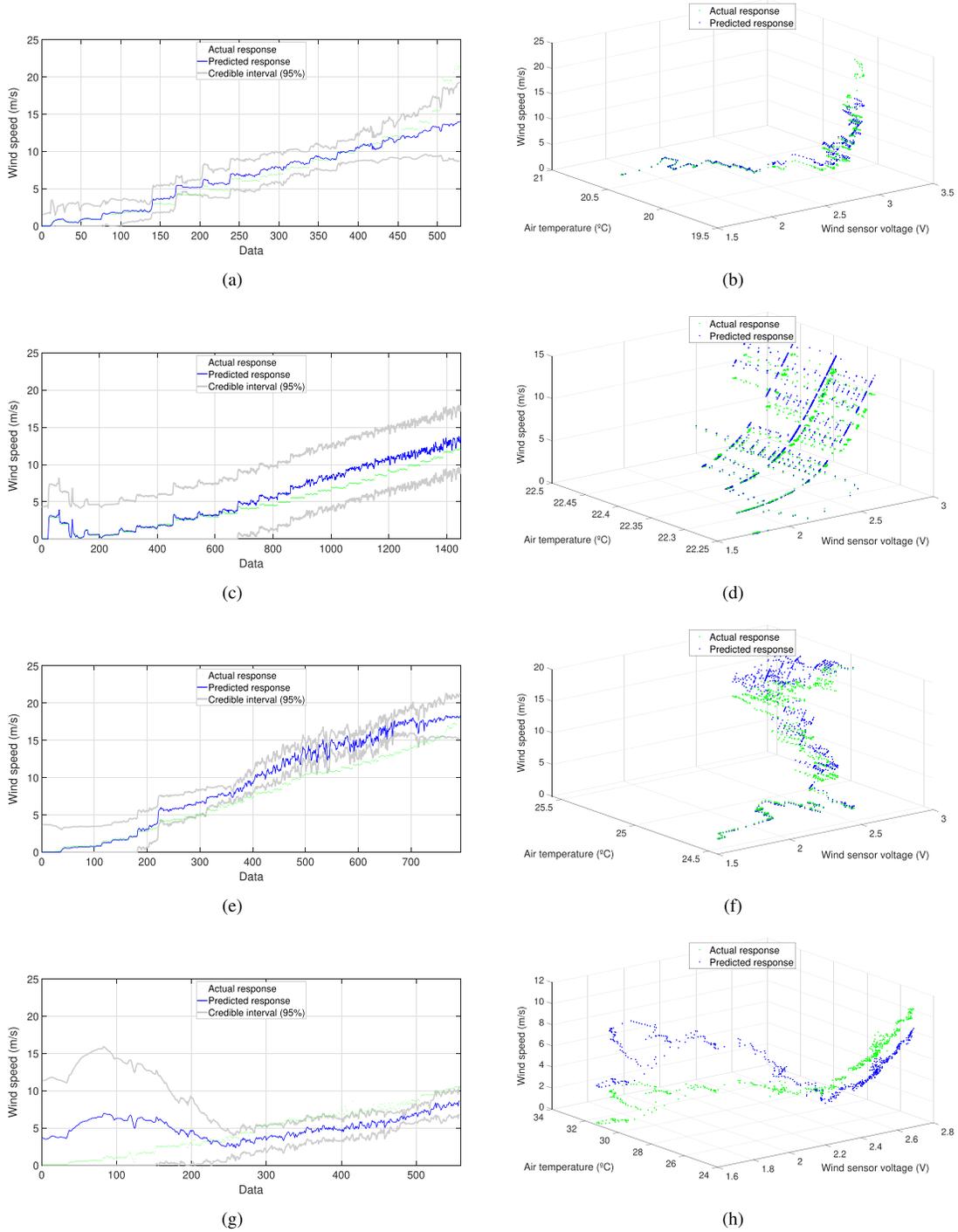

Fig. 6. Results for the GPR cross validation ("method 2"): the GPR is trained with all data set runs, except that for one particular temperature, and the resulting model is evaluated against the missing ("testing") data set. Left column shows the real wind speed and the model prediction for each point, together with its 95% credible interval. Right column shows the same data but including the raw sensor voltage and air temperature as second and third axis. Each row of images illustrates the results for a cross validation run using a different data set as "testing" data set. We show four representative such runs out of a total of seven. Notice that large errors are typically associated with large predicted uncertainty. Refer to the text for further discussion. (a) Prediction vs. ground truth (Test dataset: 20°C). (b) Prediction vs. ground truth (Test dataset: 20°). (c) Prediction vs. ground truth (Test dataset: 20°). (c) Prediction vs. ground truth (Test dataset: 22°C). (d) Prediction vs. ground truth (Test dataset: 22°C). (e) Prediction vs. ground truth (Test dataset: 24°C) (f) Prediction vs. ground truth (Test dataset: 24°C). (g) Prediction vs. ground truth (Test dataset: 30°C). (h) Prediction vs. ground truth (Test dataset: 30°C).

Table V summarizes the results for MAE, RMSE, and $R^2$, for test points of the GPR model for both types of error above. These values should be contrasted to Table IV, which shows

MAE, RMSE, and $R^2$ without adding any additional noise to measurements. As can be seen, for air temperature random errors of $\pm 0.1 \ °C$ and $\pm 0.2 \ °C$, the results of MAE, RMSE, and $R^2$ for test points are similar. For an accuracy of air temperature of $\pm 0.5 \ °C$ the variation is more remarkable

| Random error | MAE (m/s) | RMSE (m/s) | $R^2$ |
|---|---|---|---|
| $\pm$ 0.1 $°C$ | 0.1802 | 0.3057 | 0.99491 |
| $\pm$ 0.2 $°C$ | 0.2069 | 0.3411 | 0.99366 |
| $\pm$ 0.5 $°C$ | 0.3030 | 0.4855 | 0.98717 |
| $\pm$ 1 $°C$ | 0.4596 | 0.6885 | 0.97419 |

| Systematic error | MAE (m/s) | RMSE (m/s) | $R^2$ |
|---|---|---|---|
| + 0.1 $°C$ | 0.2483 | 0.4443 | 0.98950 |
| + 0.2 $°C$ | 0.3620 | 0.5658 | 0.98257 |
| + 0.5 $°C$ | 0.6622 | 0.9849 | 0.94719 |
| + 1 $°C$ | 0.9992 | 1.4516 | 0.88529 |

although it might still be acceptable, whereas for $\pm$ 1 $°C$ the error is, as expected, much greater. According to the results for systematic errors, the GPR model is more sensitive to them and consequently producing worse predictions. Systematic errors of up to + 0.2 $°C$ are acceptable.

Summarizing, attending to the results obtained, we can conclude that GPR model works with a reasonable accurate with random errors of up to $\pm$ 0.5 $°C$ or with systematic errors of up to + 0.2 $°C$.

## IV. CONCLUSION

Wind speed is a parameter hard to measure with accuracy and with low-cost devices. In this paper, the calibration of a low-cost hot-wire anemometer is proposed via machine learning techniques, attempting to solve its main drawback, namely, the loss of accuracy when air temperature changes. After comparing the performance of different regressions models, Gaussian Process Regression is the model that best fits the data and offers more precise predictions. Therefore, the problem has been addressed using Gaussian Process Regression to estimate a posterior distribution over the wind speed, given the response of a hot-wire anemometer and air temperature measurements, while also using an ultrasonic anemometer as ground truth value to rigorously evaluate the prediction error. According to the results, a low-cost hot-wire anemometer can be used, after the proposed calibration process, in different applications with reasonable accurate and over a typical range of ambient temperature.